\pdfoutput=1

\documentclass[11pt]{article}

\usepackage{EACL2023}

\usepackage{times}
\usepackage{latexsym}
\usepackage{booktabs, multirow}
\usepackage[export]{adjustbox}
\usepackage{enumitem}
\usepackage{nicematrix}

\usepackage[T1]{fontenc}

\usepackage[utf8]{inputenc}

\usepackage{microtype}

\usepackage{inconsolata}
\usepackage[font=small,skip=3pt]{caption}
\newcommand{\STAB}[1]{\begin{tabular}{@{}c@{}}#1\end{tabular}}

\newcommand{\approach}{\textsc{Forced Invalidation}}
\newcommand{\approachfi}{\textsc{FI}}
\title{Towards preserving word order importance through \\\textsc{Forced Invalidation}}

\author{Hadeel Al-Negheimish$^1$, 
 Pranava Madhyastha$^{2,1}$ \and Alessandra Russo$^1$ \\
  $^1$ Imperial College London \quad $^2$ City, University of London \\
  \texttt{\{halnegheimish,a.russo\}@imperial.ac.uk}, \\  \texttt{pranava.madhyastha@city.ac.uk}
 }

\begin{document}
\maketitle
\begin{abstract}
Large pre-trained language models such as BERT have been widely used as a framework for natural language understanding (NLU) tasks.
However, recent findings have revealed that pre-trained language models are insensitive to word order. The performance on NLU tasks remains unchanged even after randomly permuting the word of a sentence, where crucial syntactic information is destroyed.  
To help preserve the importance of word order, we propose a simple approach called \approach{} (\approachfi): forcing the model to identify permuted sequences as invalid samples. 
We perform an extensive evaluation of our approach on various English NLU and QA based tasks over BERT-based and attention-based models over word embeddings. Our experiments demonstrate that \approachfi{} significantly improves the sensitivity of the models to word order.\footnote{Our code and data for replication are available at \url{https://github.com/halnegheimish/ForcedInvalidation}}

\end{abstract}

\section{Introduction}
\label{sec:shuff-intro}
Ordering of words in a sentence is an important structural attribute for natural languages such as English, where subject-verb-object (SVO) structure is common and important to convey the meaning of the sentence. Understanding and comprehending natural language without strict adherence to a systematic ordering of words would make it an extremely challenging task. Recent work has investigated a surprising lack of sensitivity to word order information in state-of-the-art masked language models.    

Recent research has focused on the impact of word order perturbation during the evaluation of models trained on well-ordered data. 
The results show that masked language models exhibit a catastrophic lack of sensitivity to word order permutations or shuffles, even for complex tasks in which task-relevant syntactic properties are completely destroyed \cite{pham-etal-2020-order, al-negheimish-etal-2021-numerical, sinha-etal-2021-unnatural, Gupta-etal-2021-bert}. These studies show that models are still predicting the gold label for examples even after sequences have been permuted, and they also do so with high confidence \citep{sinha-etal-2021-unnatural, Gupta-etal-2021-bert}. This anomalous behaviour can potentially result in undesirable shortcuts or can cause models to fail catastrophically in simple adversarial settings. Furthermore, \citet{sinha-etal-2021-masked} study the effect of pre-training masked language models on shuffled data, and suggest that the model might simply be capturing higher-order word co-occurrence statistics, rather than uncovering sophisticated semantic and syntactic structures necessary for language understanding.

In this paper, we present a simple, yet general approach called \approach{} (FI), where we force models to explicitly identify sequence permutations (\S \ref{sec:shuff-methods}). While our proposal is extensible to multiple types of models, here we present a controlled study over masked language model-based BERT models, and attention-based models over word-embeddings (\S \ref{sec:shuff-experimental}). We present a large battery of experiments over a variety of natural language understanding tasks in the English language, including complex question answering based tasks, natural language inference based tasks and commonsense reasoning based tasks. Results show that our proposal significantly improves the sensitivity of the models to word-order information (\S \ref{sec:shuff-results}).

\section{Related Work}
\label{sec:shuff-relwork}
Recent work has proposed a few mitigation strategies for classification-type tasks: \citet{pham-etal-2020-order} proposes improving word-order sensitivity by including a precursor fine-tuning step on synthetic CoLA-like tasks before finetuning for downstream tasks. While this improves their defined word order sensitivity score, accuracy on permuted samples remains significantly above chance, making this approach unreliable. \citet{Gupta-etal-2021-bert} present three approaches: one based on entropy regularisation, another on model probabilities thresholding, and finally an approach based on augmenting additional data consisting of destructive transformations, which include 1-gram permutations. The model is modified with an additional class to identify these destructive transformations. While the first two approaches require changes to model setup, the last one is based on a set of manual heuristics. Our approach has some similarity with the latter, however, our permutations are based on the principals borrowed from the $n$-gram language modelling literature \cite{roark2007discriminative}, where the $n$-grams capture sufficient first-order statistics of the language. Our approach significantly reduces the models' reliance only on simple first-order statistics of language, and our empirical observations demonstrate the generalisability of our approach to various settings. 

\begin{figure*}[t]
\centering
\begin{minipage}{.33\textwidth}
  \centering
  \includegraphics[width=\linewidth]{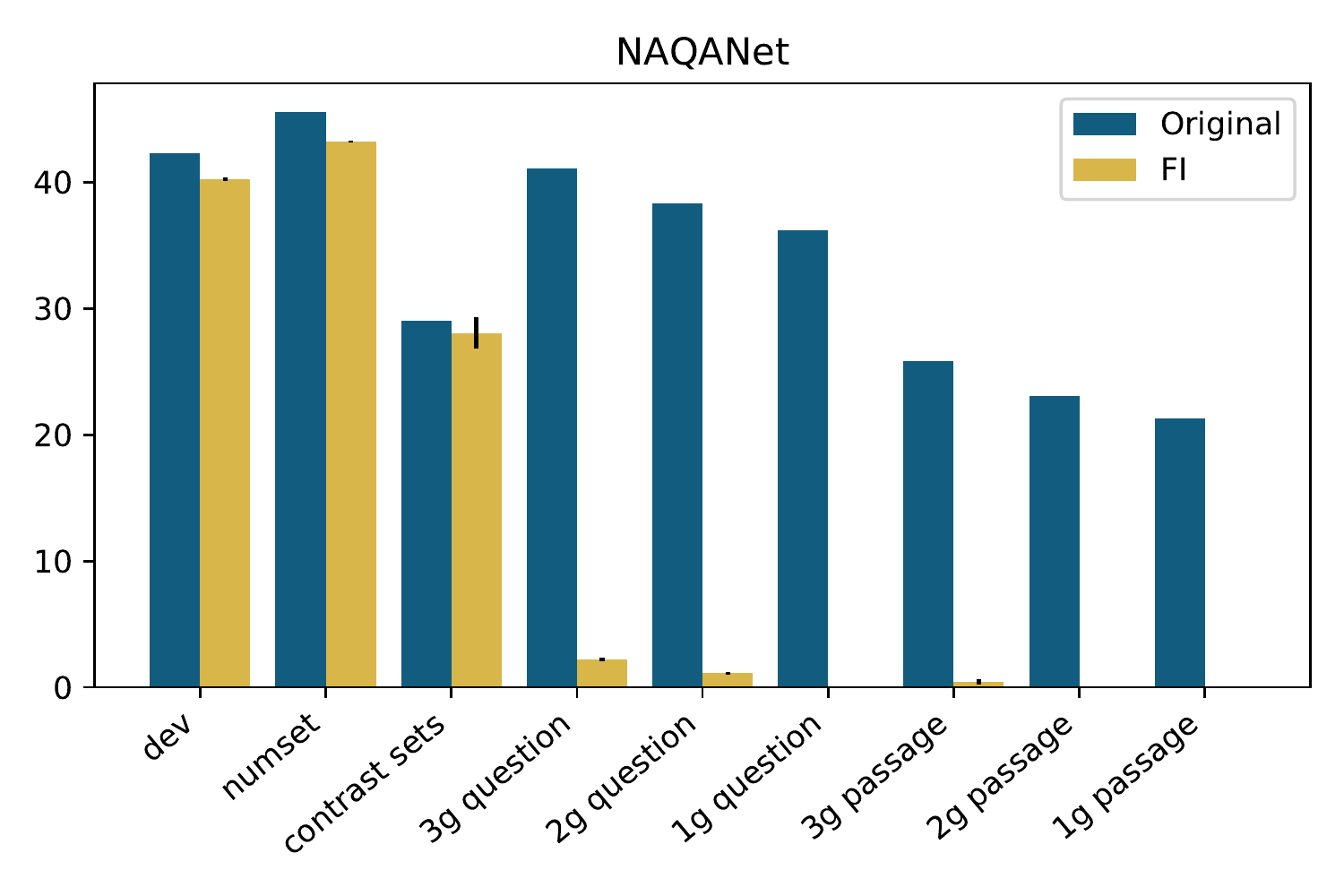}
\end{minipage}%
\begin{minipage}{.33\textwidth}
  \centering
  \includegraphics[width=\linewidth]{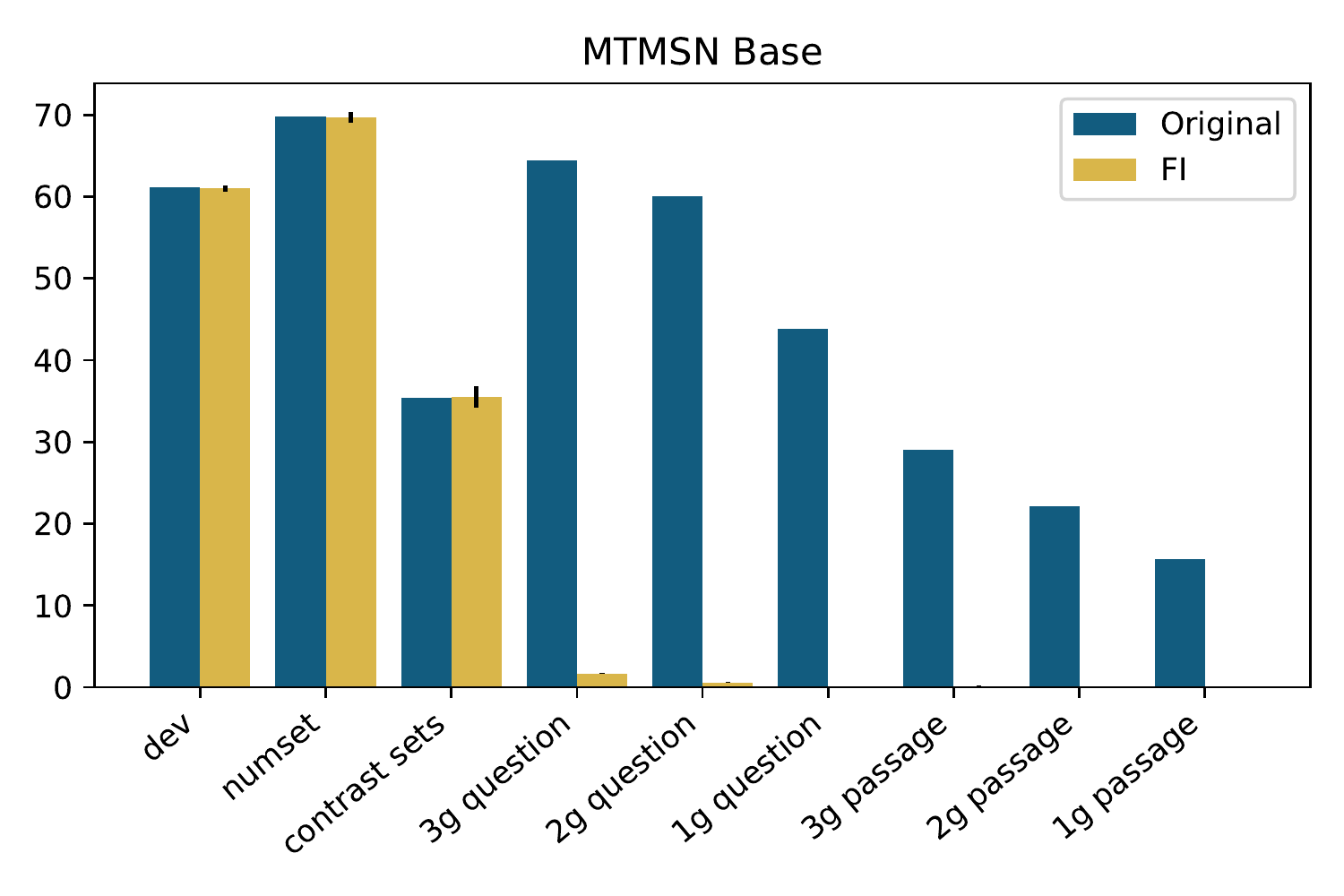}

\end{minipage}
\begin{minipage}{.33\textwidth}
  \centering
  \includegraphics[width=\linewidth]{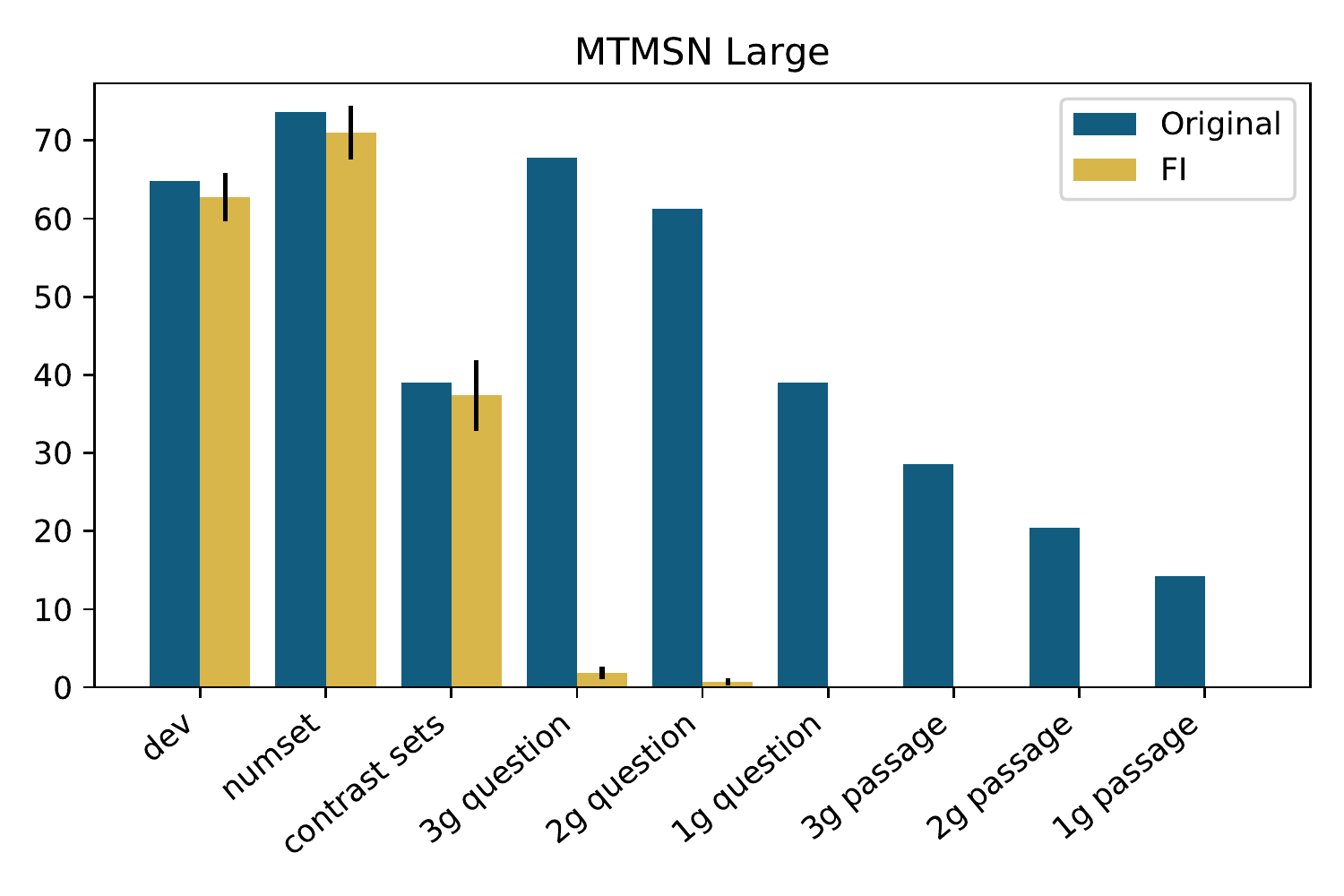}
\end{minipage}
\caption{Accuracy using exact match (EM) on the DROP QA dataset, (a) NAQANet, (b) MTMSN Base, and (c) MTMSN Large. dev, numset and contrast sets are unperturbed well-ordered datasets, while the other bars show \{1,2,3\}-gram permutations on numset. \approachfi{} models exhibit clear sensitivity to word order, as they no longer predict the original answer in most perturbed cases. FI accuracy is averaged over training with 3-random seeds, and the standard deviation is shown over the yellow bars. \approachfi{}-models are confident in identifying invalid samples.}
\label{fig:drop}
\end{figure*}

\section{Approach: \approach{}}
\label{sec:shuff-methods}
Our method is grounded on recent observations which show that masked language models and other similar models tend to exploit shortcuts based on information about distributional word vicinity information for a diverse set of natural language processing tasks~\cite{al-negheimish-etal-2021-numerical,sinha-etal-2021-masked}. These shortcuts tend to make the models less sensitive to word-order information even for tasks that require the preservation of word ordering for accurate recovery of meaning \cite{sinha-etal-2021-unnatural}. \citet{al-negheimish-etal-2021-numerical}, in particular, show the insensitivity on a variety of $n$-gram based permutations of samples, where the authors specifically experiment with \{1,2,3\}-gram permutations. Further, \{1,2,3\}-gram based permutations capture a variety of word vicinity based patterns and also capture some of the most frequently occurring unigram, bigram and trigram patterns in a variety of benchmark datasets. Based on these salient observations, for each given dataset, the \emph{\approach{}} (FI) methodology consists of the following two steps:

\begin{enumerate}[itemsep=1pt, topsep=2pt]
    \item Augmenting training data with \{1,2,3\}-gram permutation samples (sampled from trainset) labelled with \emph{invalid} as the additional label.
    \item Modifying models to account for the new label and training them in the standard setting with a combination of standard training examples and the augmented \emph{invalid} samples.
\end{enumerate}

We observe that this simple FI approach improves the sensitivity of the model to word order and also improves the robustness of the models across a variety of tasks to first-order shortcuts. In the following sections, we present a rigorous experimental study that showcases the utility of \approachfi.

\section{Experimental settings}
\label{sec:shuff-experimental}
\paragraph{Data} To generate $n$-gram permutations, we simply subsample from the training dataset, such that the ratio to the valid samples (samples with correct word order and the task-specific label) and invalid samples (samples with permuted n-grams and the \emph{invalid} label) is 1-1. The invalid samples are generated such that they contain a uniform distribution of \{1,2,3\}-gram permutations.\footnote{Input string is divided to n-grams (based on white-space), and permuted, preserving the final punctuation. The only condition is that it varies from the original string, which is a weaker constraint than previous studies that require that no n-gram stays in its original position.} Furthermore, we split the training set such that we use 90\% of the samples for training the models and the remaining 10\% is used as a development set. The development set is used to monitor training and perform early-stopping. Evaluations are done on a separate unseen task-specific validation set provided by the dataset creators.

We perform two evaluations: well-ordered and permuted. The first one is the standard evaluation of the model over the original unperturbed task-specific unseen validation set. 
For our experimental evaluations over permutations, we retain the original label of the same unseen validation set, but we permute the specific components of samples (e.g., premise or the hypothesis, question or the passage, etc.); we expect the models to reject the permuted sample instead of predicting the same ground-truth label. This setup allows us to evaluate the sensitivity of the model to word-order permutations of various degrees over the different components of the data. 

\paragraph{Models} We predominantly experiment with BERT-based models that either use BERT representations as the contextualised embeddings or classifiers that are directly trained with BERT. We also experiment with an additional simpler model that largely exploits attention over word-embeddings.\footnote{Details about training parameters can be found in the appendix~\ref{app:training}} We will expand on the specific models for each of the tasks in the following section. Results for \approachfi{} are reported over the averages and standard deviation of models trained with three random seeds.  

\section{Results and Observations}
\label{sec:shuff-results}
\subsection{Unconstrained Question Answering}

DROP \cite{Dua2019DROP} is a reading-comprehension dataset with unconstrained answers. It comprises questions that require reasoning over the content of different paragraphs. Even though this is designed to be a challenging task, 
\citet{al-negheimish-etal-2021-numerical} show that for most models, permuting questions had little impact on the model's ability to predict the correct answer for numerical reasoning questions. This was found specifically problematic as the task consists of complex questions, and permutations render the questions syntactically and semantically redundant. 
We apply \approachfi{} for two module-based models designed specifically for this task, NAQANet \citep{Dua2019DROP}, an attention-based model with GloVe-based embeddings \citep{pennington-etal-2014-glove}, which is the original model proposed by the authors of the dataset; and MTMSN \citep{hu-etal-2019-multi}, which is based on BERT-based  \citep{devlin2018bert} contextual representations. These models have separate modules targeting different kinds of reasoning, e.g. a module for counting and a module for arithmetic expressions. We augment these with an additional module to force invalidation over invalid permuted samples, a two-way classification module that learns to distinguish between permuting either the question or passage. FI models should now choose the invalid type for permuted samples, instead of giving the same answer as well-ordered samples.
Fig~\ref{fig:drop} shows a comparison of the Exact Match accuracy of NAQANet and MTMSN, between original and \approachfi{} models. We observe that question permutations have little effect on the original model, as previously noted in \citep{al-negheimish-etal-2021-numerical}. Interestingly, passage permutations can drastically reduce performance by a third. While performance degrades for passage permutations, it remains unacceptably high, as we note that DROP is an unconstrained-QA task, so the space of possible answers is large.
\approachfi{}, on the other hand, succeeds in making the model sensitive to almost all permutations. We see that this generalises across BERT$_{\small {\textsc{base}}}$ and BERT$_{\small {\textsc{large}}}$. More importantly, we observe that while the model with \approachfi{} is sensitive to word order (no longer predicting original answers for permuted examples), the model's performance on well-ordered data is largely retained. To demonstrate that \approachfi{} are correctly predicting invalid, table~\ref{table:percentage-invalid} shows the percentage of the data predicted as invalid, which is near-perfect for permuted samples.
\begin{table}[]
\adjustbox{width=\columnwidth}{%
\begin{NiceTabular}{clr|rrr|lrr}
\toprule
 && &\multicolumn{3}{c}{Part 1} &\multicolumn{3}{c}{Part 2}\\
                  &Variation & dev   & 3-gram & 2-gram & 1-gram & 3-gram & 2-gram & 1-gram \\ \midrule
\multirow{3}{*}{\STAB{\rotatebox[origin=c]{90}{UQA}}}& NAQANet            & 3.15  & 94.64 & 97.18 & 99.59 & 99.01 & 99.93 & 100.00   \\
&MTMSN Base         & 0.19  & 97.14 & 98.99 & 99.91 & 99.80  & 99.99 & 100.00   \\
&MTMSN Large        & 0.07 & 95.01 & 97.52 & 98.98 & 99.43 & 100.00   & 100.00   \\\midrule
\multirow{6}{*}{\STAB{\rotatebox[origin=c]{90}{NLI}}}&RTE                & 0.36  & 94.57 & 98.19 & 99.64 & 96.01 & 96.74 & 98.55 \\
&MNLI\_M              & 1.36  & 94.52 & 97.60  & 99.21 & 94.83 & 97.35 & 99.45 \\
&MNLI\_MM   & 1.25  & 96.20  & 98.17 & 99.49 & 94.47 & 97.43 & 99.51 \\
&ANLI1              & 0.70   & 99.80  & 100.00   & 100.00   & 95.20  & 98.10  & 99.69 \\
&ANLI2              & 0.50   & 99.90  & 100.00   & 100.00   & 93.89 & 98.50  & 99.10  \\
&ANLI3              & 1.33  & 99.83 & 99.92 & 100.00   & 94.83 & 97.33 & 98.67 \\\midrule
\multirow{1}{*}{\STAB{\rotatebox[origin=c]{90}{GA}}}&&&&&&&&\\[-2ex]
& CoLA               & 2.30   & 92.66 & 92.35 & 97.93 & -     & -     & -     \\ 

\bottomrule
\end{NiceTabular}
}
\caption{Percentage of evaluation data predicted as invalid in FI models in all of the tasks. dev is the unperturbed validation set. \{n\}-gram permutations of part 1 correspond to permutations of the question in UQA and of the premise in NLI. CoLA is made up of single sentences. All models succeed at flagging these samples as invalid.}
\label{table:percentage-invalid}
\vspace*{-4mm}
\end{table}

\begin{figure*}[t]
\centering
\begin{minipage}{.33\textwidth}
  \centering
  \includegraphics[width=\linewidth]{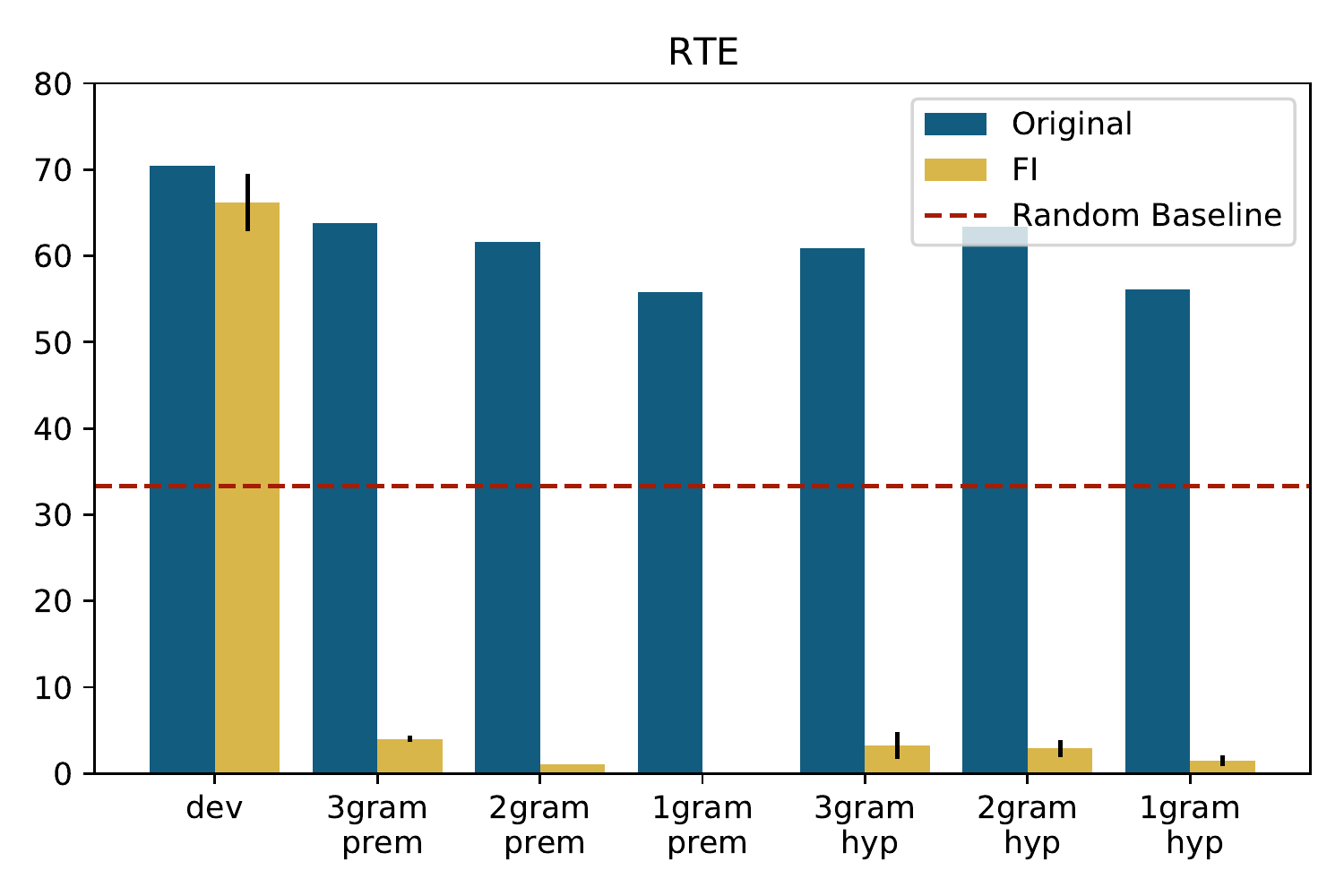}
  \includegraphics[width=\linewidth]{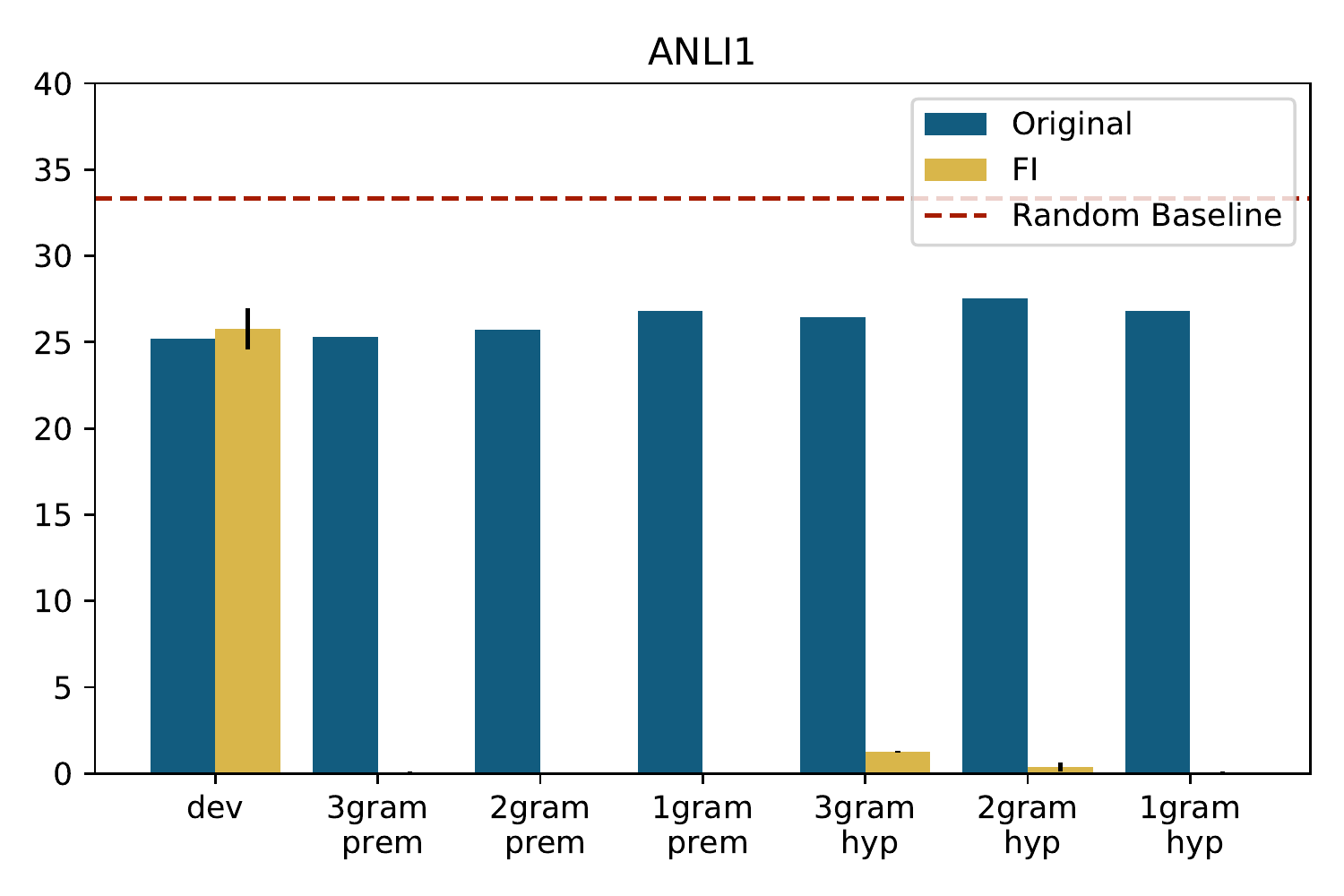}
\end{minipage}%
\begin{minipage}{.33\textwidth}
  \centering
    \includegraphics[width=\linewidth]{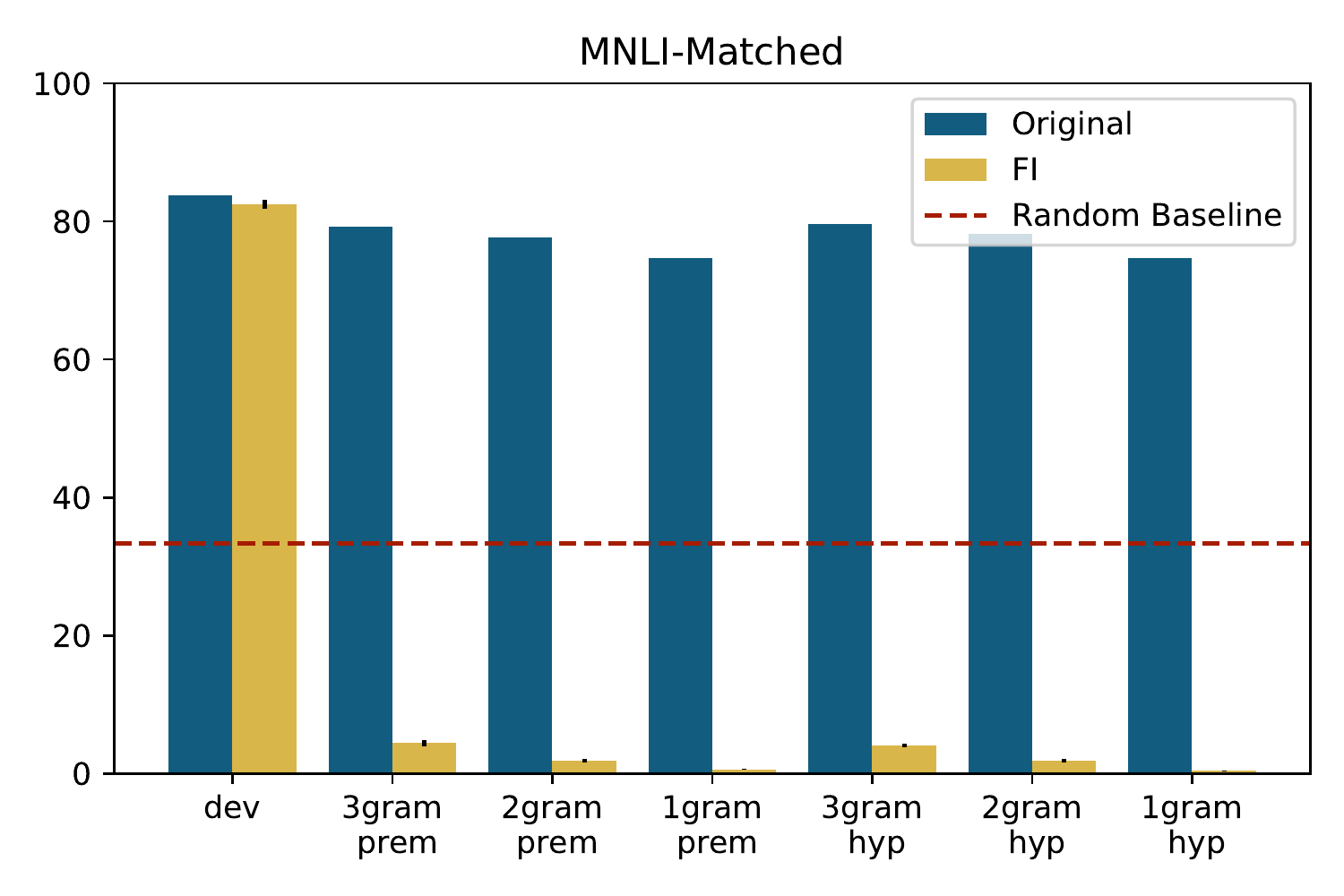}
\includegraphics[width=\linewidth]{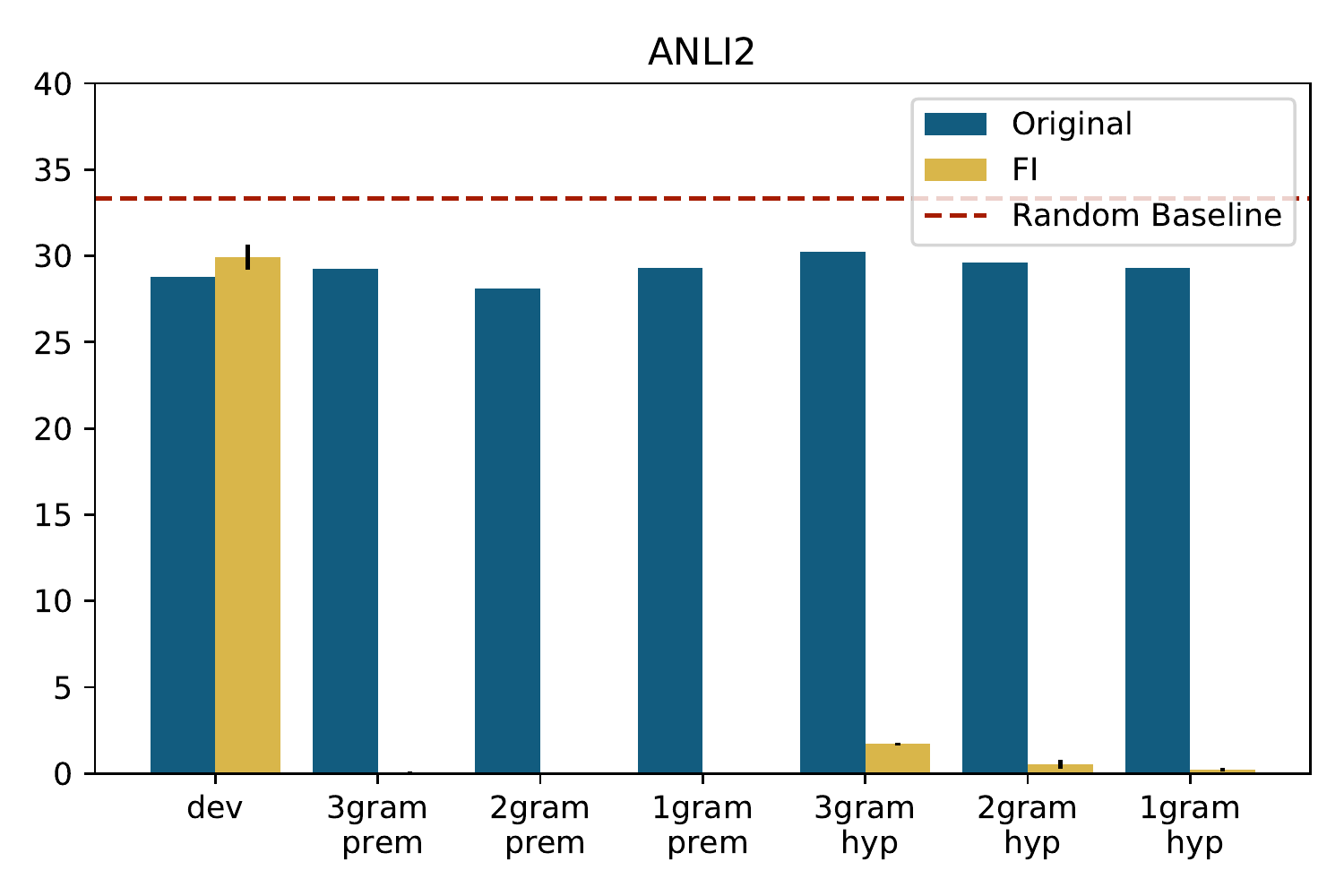}
\end{minipage}%
\begin{minipage}{.33\textwidth}
  \centering
    \includegraphics[width=\linewidth]{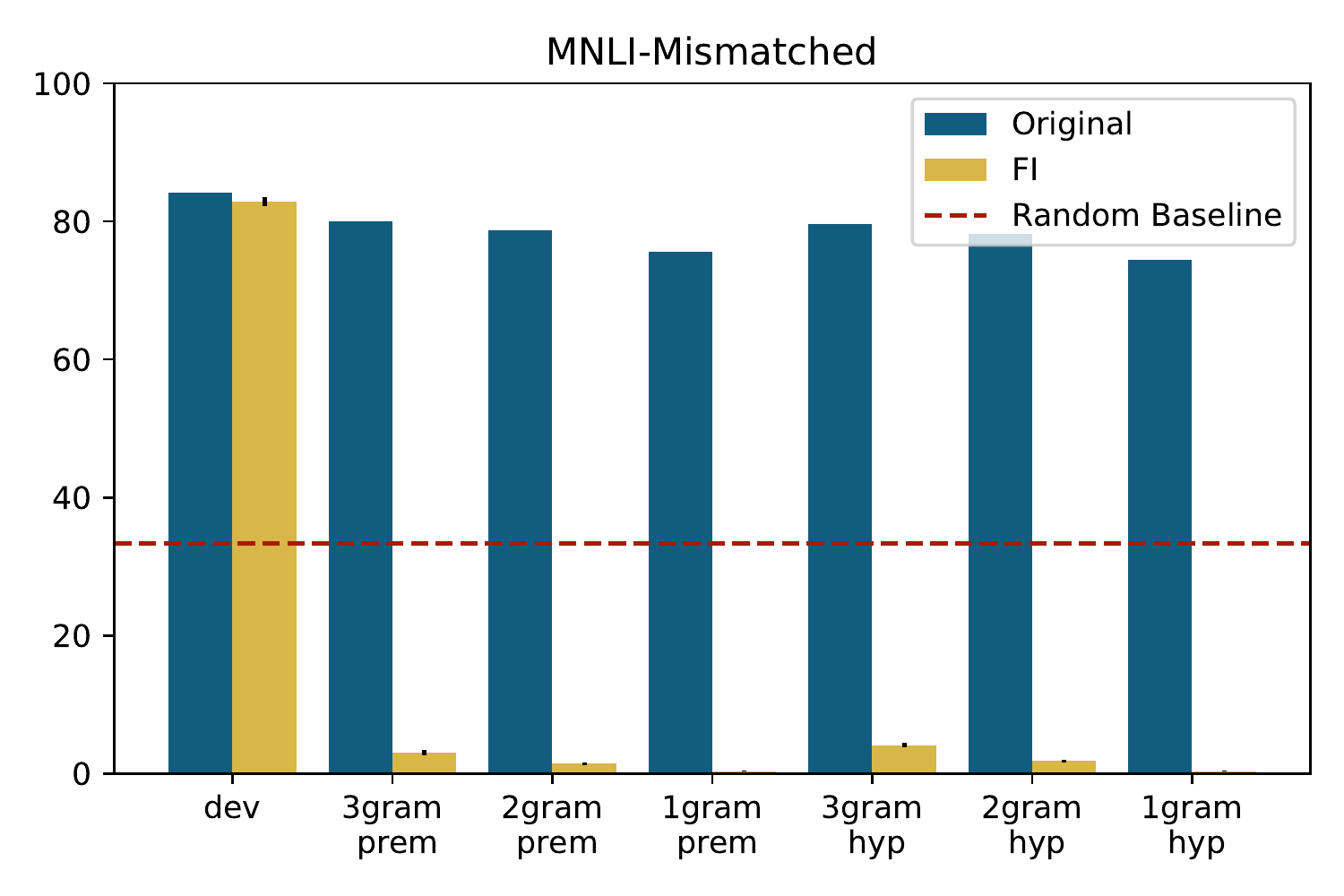}
  \includegraphics[width=\linewidth]{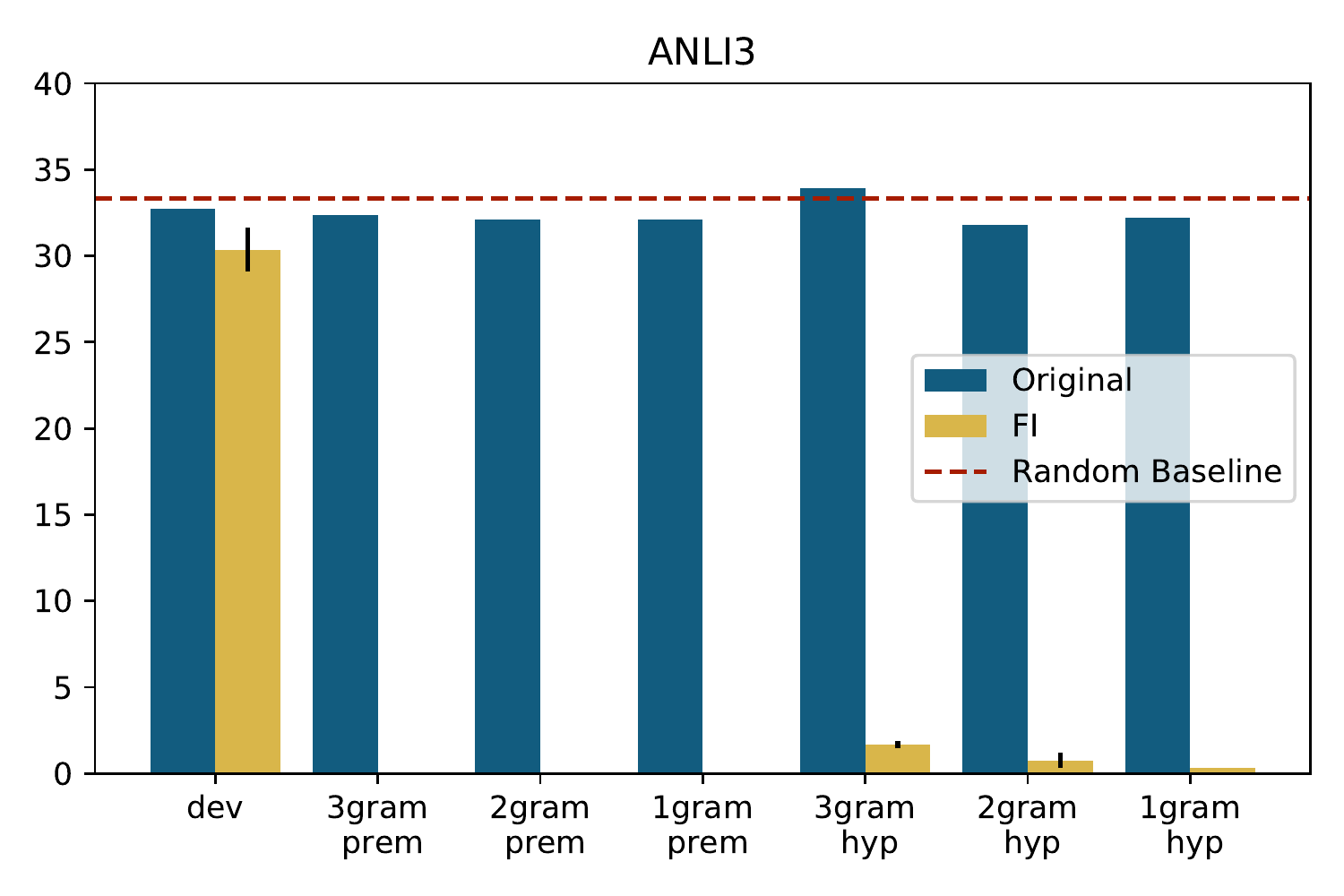}
\end{minipage}
\caption{Accuracy for NLI tasks (a) trained on RTE, (b-f) trained on MNLI data. Original models exhibit a lack of sensitivity to word order, as they have the same accuracy regardless of the n-gram permutations. FI models are able to tell apart invalid examples even in out-of-distribution ANLI data.}
\label{fig:NLI}
\vspace*{-4mm}
\end{figure*}
\subsection{Grammatical Acceptability}
\label{sec:cola}
\begin{figure}
    \centering
    \includegraphics[width=0.8\linewidth]{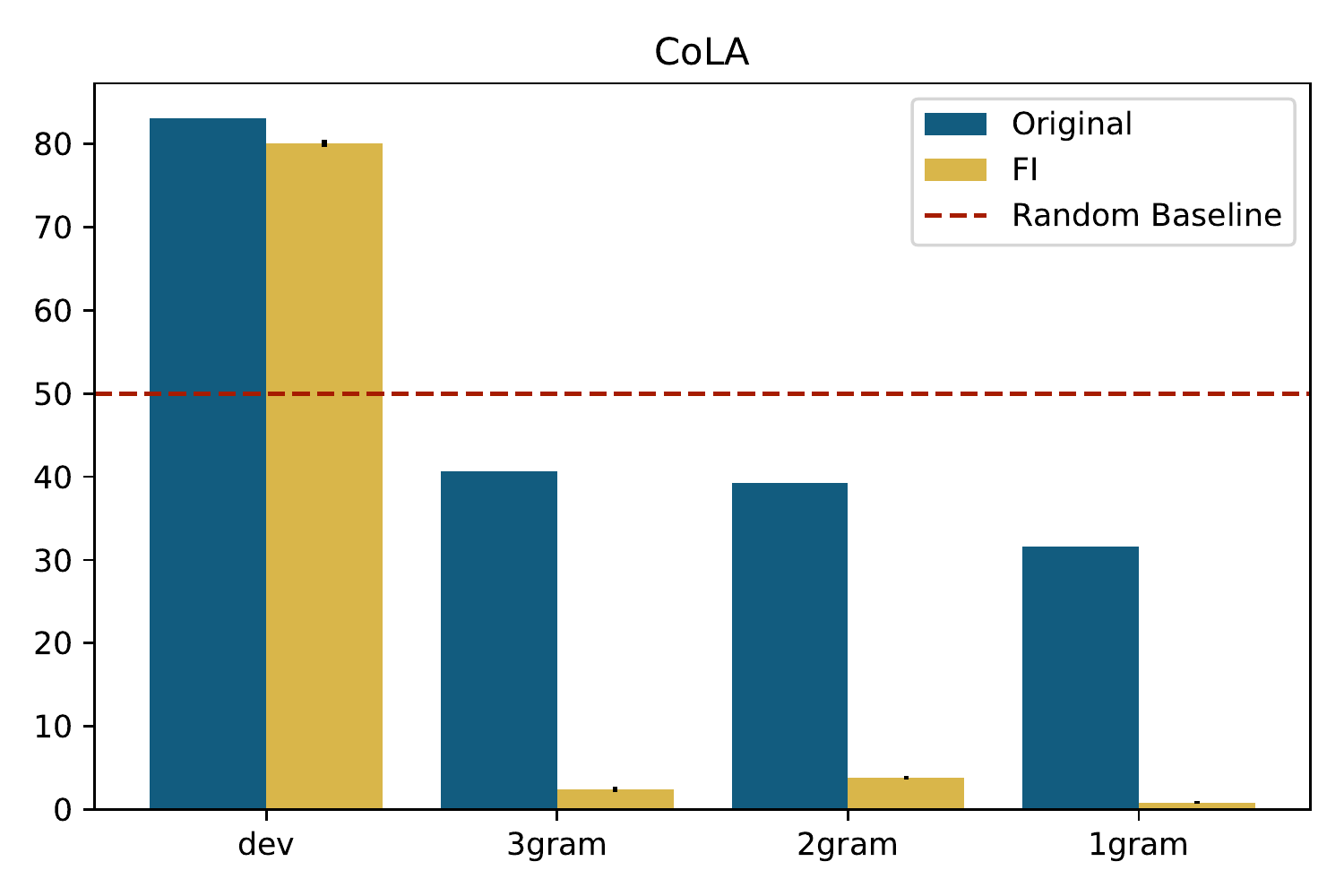}
    \caption{Accuracy of original models and \approachfi{} on CoLA grammatical acceptability task. While the original model is below chance, we note that because this task is grammar-detection, the original model should not accept any of the permuted samples. }
    \label{fig:cola}
    \vspace*{-2mm}

\end{figure}
CoLA \citep{wang-etal-2018-glue} is a task that measures models' ability to determine the grammatical acceptability of sentences. \citet{pham-etal-2020-order} show that from the GLUE benchmark \citep{wang-etal-2018-glue}, this task requires models to be most sensitive to word order. We applied \approachfi{} by extending the BERT-based classification model with another `\textit{invalid}' class label, to flag permuted sentences. Grammar and syntax have been destroyed in permuted sentences, we would expect the original model to label them as \emph{not acceptable}. However, we see in Fig~\ref{fig:cola} that it maintains significantly high accuracy for 3-gram and 2-gram permutations (note that CoLA is an imbalanced dataset (70\% acceptable)). Concretely, we observe that the standard BERT-based model labels $185/967$ 3-gram permutations, and $114/967$ 2-gram permutations as \emph{acceptable}. 
Our \approachfi{} approach significantly ameliorates this problem by increasing the sensitivity of the model to word-order permutations, as we observe in Fig~\ref{fig:cola}.
\subsection{Natural Language Inference (NLI)}
\label{sec:nli}
NLI has been one of the important testbeds of previous work studying BERT-based models and their lack of sensitivity to word-order information \citep{sinha-etal-2021-unnatural,abdou-etal-2022-word}. These studies suggest that BERT-based models almost always assign the same labels to examples with perturbed word order as well-ordered ones highlighting the lack of sensitivity to word order, and likely dependence on shallow features. Similarly to \S \ref{sec:cola}, applying \approachfi{} to an NLI BERT model was done by simply extending it with another class to represent \textit{invalid} input. We perform a battery of experiments over a variety of NLI tasks in GLUE \citep{wang-etal-2018-glue} such as RTE and MNLI \citep{williams-etal-2018-broad} tasks. \approachfi{} makes models highly sensitive to permuted sequences, as shown in Fig.~\ref{fig:NLI}. We verify in table~\ref{table:percentage-invalid} that those invalid sentences were correctly flagged as invalid. We also observe that FI-based models trained on MNLI and tested on out-of-distribution ANLI \citep{nie-etal-2020-adversarial} show extreme sensitivity to word-order perturbations; this shows that the model has learned to flag invalid input and generalise to similar unseen tasks. Additionally, we observe that FI makes the models less likely to suffer from shortcut effects, which are common in models trained for NLI tasks~\cite{mccoy-etal-2019-right}. Further experiments on Arabic NLI are presented in Appendix \ref{app:xnli}, where we demonstrate that \approachfi{} works well for other languages beyond English.
\subsubsection*{Heuristic Analysis}
\citet{mccoy-etal-2019-right} introduces an evaluation set \textit{Heuristic Analysis for NLI Systems (HANS)}, that examines surface heuristics NLI models are prone to adopting.
They show in the existence of these heuristics, models always predict entailment, as they have near-perfect accuracy for that label, but perform poorly when the actual label is non-entailment (accuracy close to 0\% in most cases). The heuristics targeted by this dataset are special cases of each other: the most general, \textit{lexical overlap} heuristic: assume that all hypotheses constructed from words in the premise are entailed. Next follows is the \textit{subsequence} heuristic: assume all hypotheses made up of contiguous subsequences of the premise are entailed. Finally, the \textit{constituent} heuristic: assume all hypotheses made up of complete subtrees of the premise's parse tree are entailed.
We expect that \approachfi-models will be more robust to these shortcuts, and find (table~\ref{table:HANS-results}) that this is indeed the case for lexical-overlap and the more challenging  subsequence heuristics, substantially improving non-entailment performance, indicating that the models are no longer strictly biased with the presence of these heuristics. We note a surprising result of a drop in accuracy for entailed lexical overlap samples, which could be caused by the model no longer taking that shortcut, and warrants additional investigation in future work.
\begin{table}[]
\adjustbox{width=\columnwidth}{%
\begin{tabular}{@{}lrp{1.5cm}rr@{}}
\toprule
                                   &                & \textbf{{Lexical Overlap}} & \textbf{Subsequence} & \textbf{Constituent} \\ \midrule
\multirow{2}{*}{\textbf{Original}} & Entailment     & 97.76                    & 99.92                & 100.00                  \\
                                   & Non-Entailment & 14.74                    & 0.58                 & 2.66                 \\\midrule
\multirow{2}{*}{\textbf{FI}}       & Entailment     & 80.94                    & 99.56                & 99.60                \\
                                   & Non-Entailment & 64.56                    & 38.14                & 11.48                \\ \bottomrule
\end{tabular}
}
\caption{Comparison of BERT finetuned on MNLI with and without \approachfi, \approachfi{} makes the model more robust to the syntactic heuristics presented in \citep{mccoy-etal-2019-right}}
\label{table:HANS-results}
\end{table}

\begin{figure}
    \centering
    \includegraphics[width=0.8\linewidth]{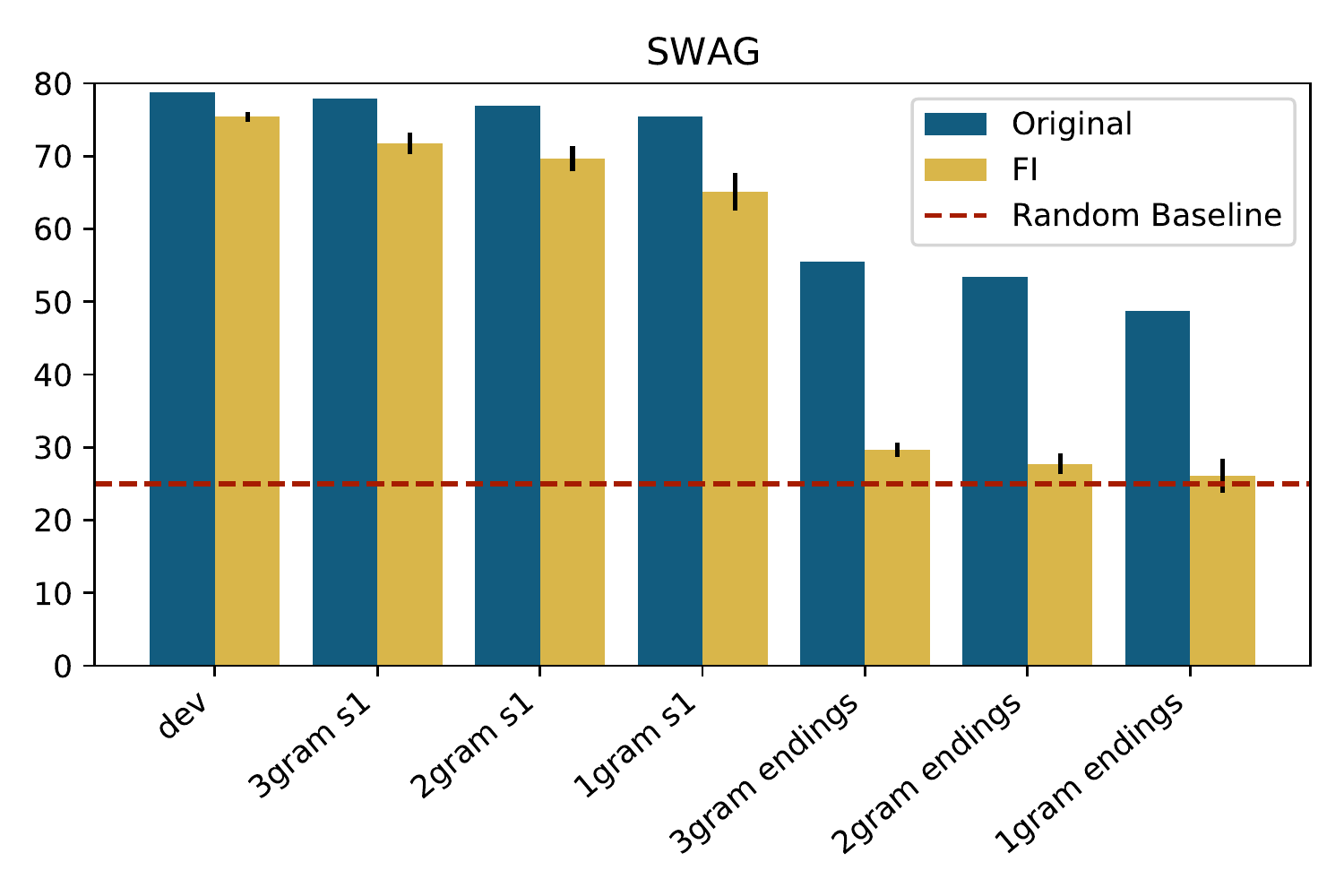}
    \caption{Accuracy of original models and \approachfi{}  on SWAG multiple-choice commonsense reasoning task, both are presented with the same four answer choices, \approachfi{} enforces sensitivity to word order, especially on the endings.}
    \label{fig:swag}
    \vspace*{-3mm}
\end{figure}

\subsection{Multiple Choice Commonsense Reasoning}
\label{sec:swag} 
SWAG \citep{zellers-etal-2018-swag} is a commonsense and grounded reasoning task that requires choosing between different possible ending scenarios given some context, where the context comprises of a primary sentence followed by the initial set of words for the following sentence.  
The model is trained to predict the most likely ending scenario given context and a list of four ending scenarios. To train our \approach{} model, we add an additional answer choice \texttt{`is invalid.'} to the data and perform $n$-gram based permutation of the primary sentence or over each of the endings. Nothing is changed in the architecture of the model. For evaluation, we maintain the same datasets for the Original and \approachfi{} models, so they only contain the same number of answer choices, without the \texttt{`invalid'} choice. Ideally, the models should achieve random performance in the permuted cases. We present our results in Fig.~\ref{fig:swag}; where firstly we observe that \approachfi{} does not seem to reliably affect the model's sensitivity to word order perturbations with all the combinations. We specifically observe that $n$-gram permutations of primary sentence have little impact on the performance of \approachfi-models and they seem to be less sensitive to word-order perturbations than expected. However, $n$-gram permutations of endings result in \approachfi-models obtaining near-random performance as expected.
We investigated the cause of this anomaly and observed that SWAG dataset has a prominent problem, in that, the primary sentence for a majority of cases is almost irrelevant to the model. A model trained on SWAG dataset is able to predict the correct answer for 60\% of the examples without having access to the primary sentence. 

\section{Conclusion}
\label{sec:shuff-conc}
In this paper, we presented a simple yet general technique called \approach{}, that significantly improves the sensitivity of models towards word order information. Our methodology requires minimal changes to the model and is sample efficient and drastically increases the sensitivity of the models to permutations of word order for a variety of tasks. We present a focused empirical validation of our methodology to showcase its generalisability. While in this paper we have focused on masked language models and attention-based models over word embeddings, we expect \approachfi{} to generalise for other modelling setups such as RNN-based and CNN-based models and leave it as future work. We anticipate that this approach will also serve as a solution for other undesired behaviours in the model by explicitly invalidating such behaviours. We leave this as future work.


\section*{Limitations}
While our empirical results showcase the effectiveness of FI and increase models' sensitivity to word order, the causal mechanisms are not currently obvious. It is not clear whether or not positional encodings are reflecting this change. Like previous work, our observations are additionally only restricted currently to English and Arabic (appendix \ref{app:xnli}), further experiments are required to establish the problems relating to word order sensitivity and the utility of \approachfi{} for other languages. 

\section*{Acknowledgements}
This research has been supported by a PhD scholarship from King Saud University. We thank our anonymous reviewers for their constructive feedback.
\bibliography{anthology,custom}
\bibliographystyle{acl_natbib}

\appendix
\section{Training Details}
\label{app:training}

We use BERT$_{\small {\textsc{base}}}$ models with a classification head on top (BERTForSequenceClassification \citep{wolf-etal-2020-transformers}) for NLI and CoLA. It was trained for 5 epochs, batch size 16, learning rate 2e-5. 
We use a BERT$_{\small {\textsc{base}}}$ model \citep{wolf-etal-2020-transformers}, for SWAG. It was trained for 3 epochs, batch size 16, learning rate 5e-5. All of the above are done using google colab with high-ram.
MTMSN is based on the published codebase \citep{hu-etal-2019-multi}, and we use the same parameters to train, namely: BERT$_{base}$: batch size 24, 5 epochs and  learning rate 3e-5, BERT$_{large}$: batch size 12, 10 epochs and  learning rate 3e-5. Training was done on four RTX6000 GPUs with 24GB of RAM each for BERT$_{\small {\textsc{large}}}$, and a single one was used for BERT$_{\small {\textsc{base}}}$. 
Models are trained for both settings, original and \approachfi, to provide a fair comparison. 

\section{Dataset Statistics}
As described before, we filter out examples with sentences containing less than three words, such that we can generate at least 1 3-gram shuffle. Table~\ref{table:dataset-stats} describes the tasks' validation sets before and after filtration.

\begin{table}[h]
\adjustbox{width=\columnwidth}{%
\begin{tabular}{@{}lllll@{}}
\toprule
            & Original & Used  & p1 \#words & p2  \#words \\ \midrule
DROP numset & 6848     & 6848  & 11         & 182         \\
RTE         & 277      & 276   & 31         & 8           \\
MNLI        & 9815     & 9289  & 16         & 9           \\
MNLI-mm     & 9832     & 9551  & 17         & 10          \\
ANLI1       & 1000     & 999   & 54         & 10          \\
ANLI2       & 1000     & 999   & 54         & 9           \\
ANLI3       & 1200     & 1199  & 52         & 9           \\
CoLA        & 1043     & 967   & 7          & -           \\
SWAG        & 20006    & 19352 & 11         & 8           \\ \bottomrule
\end{tabular}
}
\caption{Statistics of the validation set of datasets used for evaluation, including the original number of examples, after filtration, and median number of words for the first and second components. }
\label{table:dataset-stats}
\end{table}

Dataset licenses are mentioned in table~\ref{table:dataset-licenses}:
\begin{table}[h]
\centering
\adjustbox{width=0.5\columnwidth}{%
\begin{tabular}{@{}ll@{}}
\toprule
            & License \\ \midrule
DROP numset &    CC BY 4.0     \\
RTE         &    Unknown        \\
MNLI        &     OANC        \\
MNLI-mm     &      OANC \\
ANLI1       &    CC BY-NC 4.0     \\
ANLI2       &   CC BY-NC 4.0    \\
ANLI3       &   CC BY-NC 4.0        \\
CoLA        &    Unknown        \\
SWAG        &    MIT license         \\ \bottomrule
\end{tabular}
}
\caption{Artifact licenses for the datasets used.}
\label{table:dataset-licenses}
\end{table}

\section{\approachfi{} as a precursor to downstream task finetuning}

Inspired by \citep{pham-etal-2020-order}, we first perform \approachfi{} on BERT to solely categorise valid and invalid samples for sentences that are sampled from Wikipedia. We replace the standard BERT-based contextual representations in MTMSN with FI-BERT based contextual embeddings, to see if it helps it become more sensitive to word order in the downstream DROP task. 
This did not show an improvement, however, where permuted examples are still predicted the same as well-ordered ones. This indicates that having an explicit way to flag invalid examples is helpful to the models.

\section{\approach{} with Other Languages}
\label{app:xnli}
\begin{figure}
    \centering
    \includegraphics[width=0.8\linewidth]{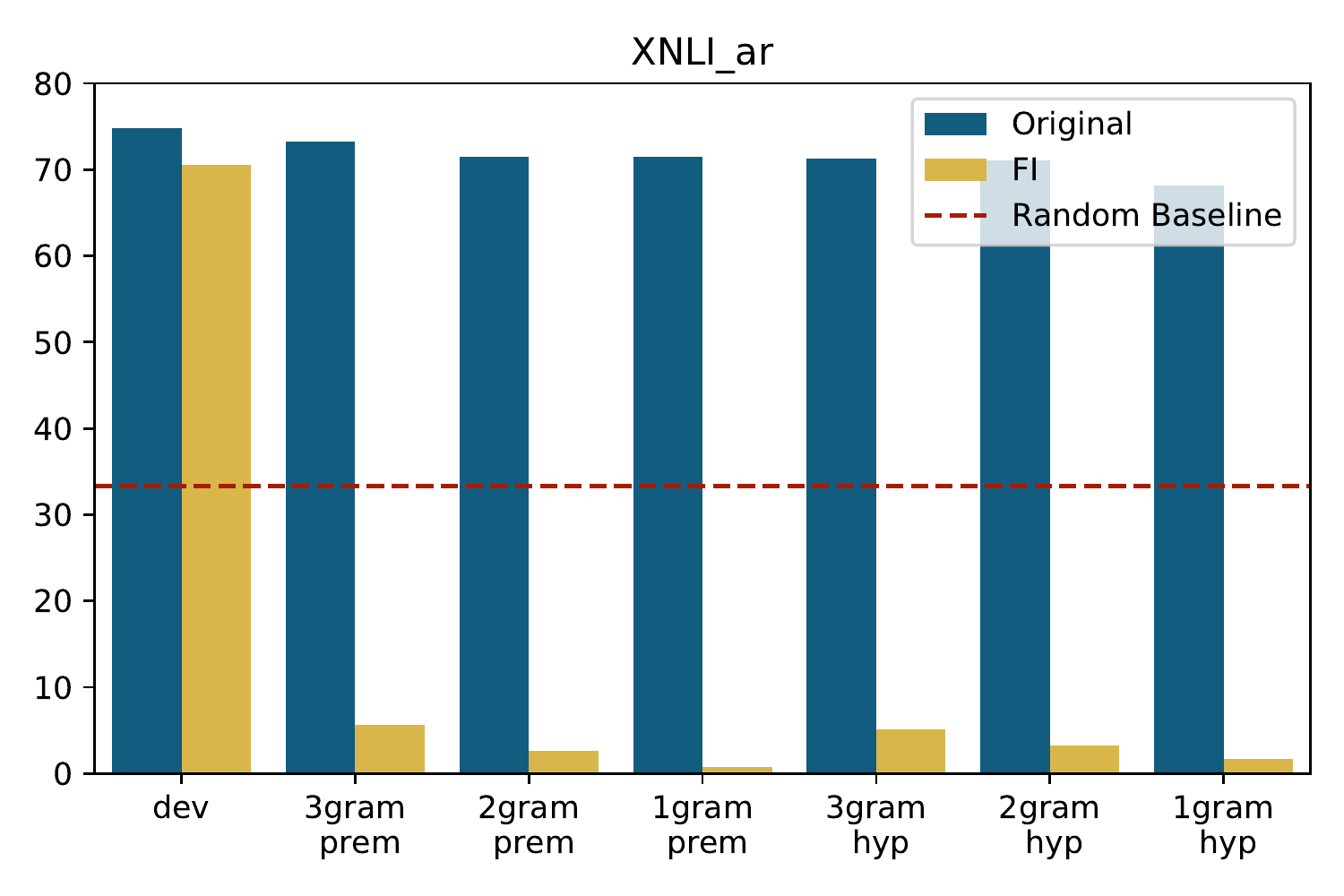}
    \caption{Accuracy of original models and \approachfi{}  on an Arabic NLI task. Once more, we see that the original models are insensitive to word order even in Arabic, while \approachfi{} models learn to flag invalid samples.}
    \label{fig:xnli_ar}
\end{figure}

To establish the generalizability of this approach to other languages, we applied \approachfi on an Arabic NLI task. We used the Arabic split of the XNLI dataset \citep{conneau-etal-2018-xnli} to finetune the ArBERT model \citep{abdul-mageed-etal-2021-arbert}, and compare the original training setup with \approach{}. Figure~\ref{fig:xnli_ar} shows that this approach successfully preserves the importance of word order beyond the English language.
\end{document}